\documentclass{article}
\usepackage{spconf,amsmath,graphicx}

\usepackage{multirow}
\usepackage{mathtools}
\usepackage{amssymb}
\usepackage{booktabs}
\usepackage{balance}
\usepackage[skip=1.5pt]{caption}
\title{MONet: Multi-scale Overlap Network for Duplication Detection in Biomedical Images}
\name{Ekraam Sabir\textsuperscript{*}, Soumyaroop Nandi\textsuperscript{*}, Wael AbdAlmageed\textsuperscript{*\dag}, Prem Natarajan\textsuperscript{*}}
\address{\textsuperscript{*}USC Information Sciences Institute, Marina del Rey, CA, USA \\ \textsuperscript{\dag}Department of Electrical and Computer Engineering, USC, Los Angeles, USA}
\begin{document}
%
\maketitle
\begin{abstract}
Manipulation of biomedical images to misrepresent experimental results has plagued the biomedical community for a while. Recent interest in the problem led to the curation of a dataset and associated tasks to promote the development of biomedical forensic methods. Of these, the largest manipulation detection task focuses on the detection of duplicated regions between images. Traditional computer-vision based forensic models trained on natural images are not designed to overcome the challenges presented by biomedical images. We propose a multi-scale overlap detection model to detect duplicated image regions. Our model is structured to find duplication hierarchically, so as to reduce the number of patch operations. It achieves state-of-the-art performance overall and on multiple biomedical image categories. 
\end{abstract}
\section{Introduction}\label{sec:intro}

Advancements in multimedia technology has enabled the proliferation of digitally manipulated misinformation. Prevalent manifestions of misinformation include fake news, digitally manipulated images, deepfake videos and more. Efforts towards the development of automated detection methods for fake news \cite{sabir_deep_2018,muller2020multimodal,sabir2021meg}, natural-image forensics \cite{Wu_2018_ECCV,Wu_2019_CVPR,wang2022objectformer} and deepfakes \cite{sabir2019recurrent,masi2020two,Dong_2022_CVPR} have gained traction. However, an important yet almost neglected field is that of biomedical image forensics. Misrepresentation of experimental results by manipulating biomedical images has been an issue of concern for a while in the biomedical community \cite{bik2016prevalence}. Unlike natural images, biomedical images often contain arbitrary patterns with no semantic context which allows for these manipulations to go undetected during the peer review process. Investigative or follow-up research can lead to the discovery of such manipulations which consequently leads to retractions \cite{bik2018analysis}. However the entire discovery process is a loss of time and money \cite{stern2014financial}. 

\begin{figure}
    \centering
    \includegraphics[width=0.85\linewidth]{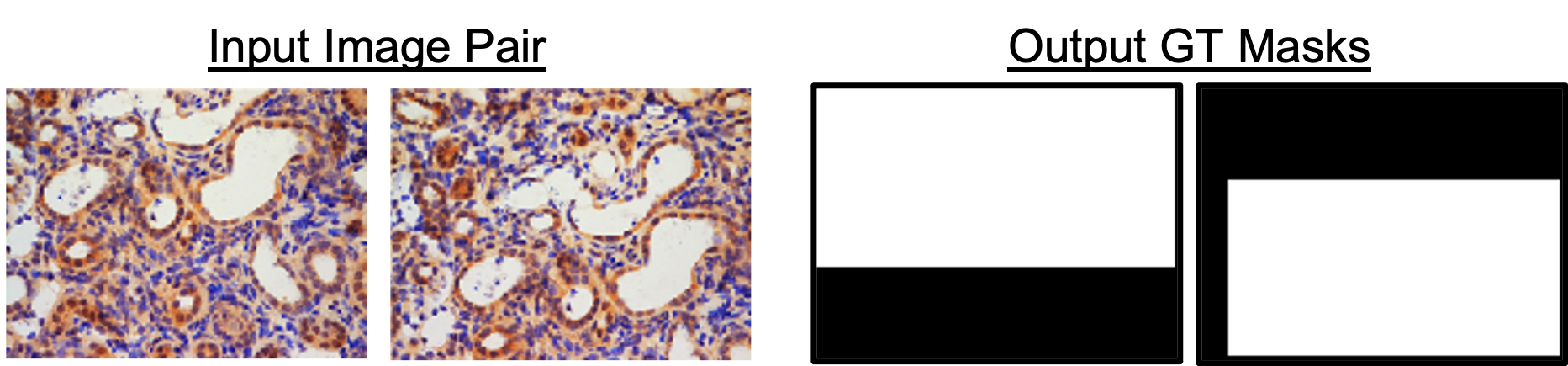}
    \caption{An image pair sample from the EDD task in BioFors.}
    \label{fig:teaser}
\end{figure}

Recently, a new biomedical image forensics dataset (BioFors) \cite{sabir2021biofors} was released to promote the development of automated detection methods. The dataset comprises images belonging to four categories collected from biomedical research documents. The paper also proposed three biomedical forensic tasks to overcome the lack of structured problem definitions in literature. There are three manipulation detection tasks described in \cite{sabir2021biofors} -- external duplication detection (EDD), internal duplication detection (IDD) and cut/sharp transition detection (CSTD). These tasks together cover popular forms of semantic and digital manipulations found in biomedical literature. Of these problems, we focus on the largest task involving the detection of duplicated regions between images a.k.a the external duplication detection (EDD) task. The provenace of manipulations in this task is not known i.e. the images could be spliced, cropped with an overlap from a larger image or simply reused. Figure \ref{fig:teaser} shows a manipulated sample under the EDD task from \cite{sabir2021biofors}.

Related research areas of image matching and splicing detection are matured. However, as shown in \cite{sabir2021biofors}, traditional computer vision methods trained on natural images are not suitable for biomedical forensics. Difficulties in detecting keypoints from biomedical images limit keypoint-descriptor based methods \cite{lowe2004distinctive,rublee2011orb,calonder2010brief}. Use of coarse feature maps limits the detail in deep-learning based splicing detection methods \cite{wu2017deep,Wu_2018_ECCV} and dense matching of features is computationally expensive \cite{cozzolino2015efficient}. To overcome these challenges, we propose a multi-scale overlap detection network (MONet) that recursively finds overlap between patches to locate duplicated image regions. Recursive overlap detection is performed at multiple scales in an hierarchical manner from large to small image patches. Our model increases the matching detail from coarse to refined feature maps in a top-down approach, while simultaneously reducing the computational burden by making fewer patch-comparisons. 

\begin{figure*}[t!]
    \centering
    \includegraphics[width=0.85\textwidth]{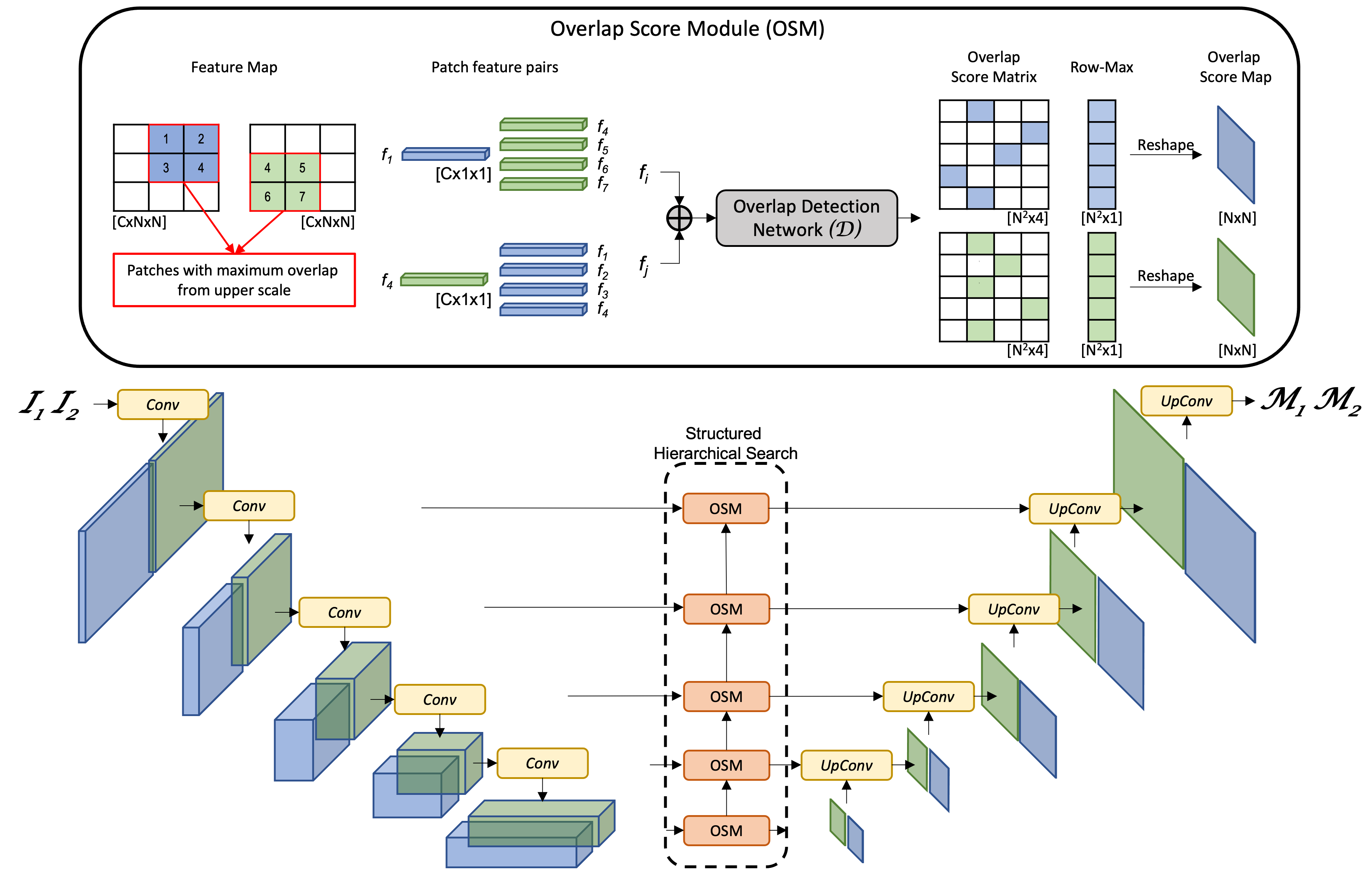}
    \caption{Illustration of MONet. The top shows details of overlap score module (OSM) and the bottom shows overall architecture.}
    \label{fig:model}
\end{figure*}
\section{Proposed Method}\label{sec:method}

The EDD task is structured to locate duplicated regions between image pairs. Given two input images ${I_{1}, I_{2} \in \mathbb{R}^{H \times W \times 3}}$ the objective of the EDD task is to predict two binary masks $M_{1}, M_{2} \in \mathbb{R}^{H \times W \times 1}$, highlighting the duplicated image region as shown in Figure \ref{fig:teaser}. Our model comprises convolutional encoding and decoding operations at multiple patch scales as shown in Figure \ref{fig:model}. The overlap-score module (OSM) performs overlap detection at each scale. The search for overlapping image regions is structured hierarchically across scales by linked OSMs.

\subsection{Architecture Overview} The general structure of our model resembles a U-Net \cite{ronneberger2015u} with a series of convolutional encoders at multiple scales $s \in \{1,2,3,4,5\}$ that produce feature maps $F \in \mathbb{R}^{N_{h} \times N_{w} \times C_{s}}$. For notational convenience we consider that images are square with $N \times N \times 3$ dimension. Consequently, the dimension of encoder feature maps at each scale is $\frac{N}{2^{s}} \times \frac{N}{2^{s}}$. The upsampling involves a series of convolutional decoders that produce feature maps at corresponding scales to that of the encoder. The final output of the decoder produces a pair of binary prediction masks. To find duplicated regions between images, we measure overlap between patches of two images at each scale within overlap-score modules (OSMs) in a top-down hierarchy. The maximum overlap score of a patch indicates the confidence with which all or a part of it is considered to have been repeated in the other image. A higher score indicates full or substantial repetition, while a low score represents negligible or no repetition. To minimize the number of patch comparisons, patch pairs in $I_{1}$ and $I_{2}$ with maximum overlap at the current scale $s$ are used to guide the search among sub-patches at the lower scale $s-1$. 

\begin{table}[tbh]
\small
    \centering
    \begin{tabular}{c|c|c|c}
    \hline
         \multirow{2}{*}{\textbf{Scale}} & \textbf{Patch} & \textbf{Naive} & \multirow{2}{*}{\textbf{Ours}} \\
         & \textbf{Dimension} & \textbf{Comparisons} & \\
         \hline
         1 & 2x2 & $\sim$ 268.43M  & $\sim$ ~131K \\
         2 & 4x4 & $\sim$ 16.77M & 32,768 \\
         3 & 8x8 & $\sim$ 1.04M & 8,192 \\
         4 & 16x16 & 65,536 & 2,048 \\
         5 & 32x32 & 4,096 & 4,096 \\
         \hline
    \end{tabular}
    \caption{Number of patch comparisons at each scale.}
    \label{tab:patch_comp}
\end{table}

\begin{table*}[t!]
\small
    \centering
    \begin{tabular}{lcccccccccc}
    \toprule
         \multirow{3}{*}{\textbf{Method}} &  \multicolumn{2}{c}{\textbf{Microscopy}} & \multicolumn{2}{c}{\textbf{Blot/Gel}} & \multicolumn{2}{c}{\textbf{Macroscopy}} & \multicolumn{2}{c}{\textbf{FACS}} & \multicolumn{2}{c}{\textbf{Combined}} \\
         \cmidrule(lr){2-3}\cmidrule(lr){4-5}\cmidrule(lr){6-7}\cmidrule(lr){8-9}\cmidrule(lr){10-11}
         & Image & Pixel & Image & Pixel & Image & Pixel & Image & Pixel & Image & Pixel \\
         \cmidrule(r){1-1}\cmidrule(l){2-11}
         SIFT \cite{lowe2004distinctive} & 0.180 & 0.146 & 0.113 & 0.148 & 0.130 & 0.194 & 0.11 & 0.073 & 0.142 & 0.132 \\
         ORB \cite{rublee2011orb} & 0.319 & 0.342 & 0.087 & 0.127 & 0.126 & 0.226 & 0.269 & 0.187 & 0.207 & 0.252 \\
         BRIEF \cite{calonder2010brief} & 0.275 & 0.277 & 0.058 & 0.102 & 0.135 & 0.169 & 0.244 & 0.188 & 0.180 & 0.202 \\
         DF - ZM \cite{cozzolino2015efficient} & \textbf{0.422} & 0.425 & 0.161 & 0.192 & 0.285 & 0.256 & \textbf{0.540} & \textbf{0.504} & 0.278 & 0.324 \\
         DMVN \cite{wu2017deep} & 0.242 & 0.342 & 0.261 & 0.430 & 0.185 & 0.238 & 0.164 & 0.282 & 0.244 & 0.310 \\
         \cmidrule(r){1-1}\cmidrule(l){2-11}
         \textbf{Ours - regular margin loss} & 0.398 & \textbf{0.435} & 0.507 & \textbf{0.520} & 0.221 & 0.262 & 0.313 & 0.356 & \textbf{0.410} & \textbf{0.438} \\
         \textbf{Ours - flexible margin loss} & 0.346 & 0.386 & \textbf{0.520} & \textbf{0.520} & \textbf{0.309} & \textbf{0.281} & 0.256 & 0.336 & 0.398 & 0.410 \\
         \bottomrule
    \end{tabular}
    \caption{MCC scores on external duplication detection (EDD) task in BioFors across image categories.}
    \label{tab:result}
\end{table*}

\subsection{Overlap-Score Module (OSM)} The purpose of the OSM module is to predict two overlap score maps at each scale corresponding to the feature maps. Deconvolution layers upsample the overlap score maps sequentially to produce binary output masks. Score maps from previous and current scale are concatenated for upsampling. Overlap scores are produced by an overlap detection network $\mathcal{D}$ which takes as input two patch feature vectors (one from each image). It is trained on patch feature triplets (anchor, overlapping and non-overlapping patches) generated from synthetic data at each scale. We consider a feature map $F$ at scale $s$, to be composed of a grid of patch feature vectors $f \in \mathbb{R}^{1 \times 1 \times C_{s}}$, such that each feature vector represents a patch of dimension $d_{s} \times d_{s}$ in the input image, where $d_{s}=\frac{N}{2^{s}}$. While the convolutional receptive field of a feature vector $f$ is larger than the patch dimension $d_{s}$ at any given scale, we implicitly limit the scope of each feature vector to its patch dimensions when measuring overlap. The overlap score map, is indexed similar to a feature map $F$. The score at each index represents the maximum overlap found for that patch feature vector when compared to vectors from the other image. 

\subsection{Structured Hierarchical Search} The OSMs are structurally linked from higher to lower scale such that patch comparisons can be made hierarchically.Sub-patches of a patch with maximum overlap at a higher scale, are candidate patches for overlap detection at a lower scale. Since, the spatial dimension of each feature map gets halved at each scale, a feature vector $f$ at a higher scale overlaps with four feature vectors at the immediate lower scale. This observation is useful in limiting the number of patch comparisons to to be made at a lower scale. For two patches with maximum overlap at a higher scale, each of their four sub-patches are compared only with each other. At the largest scale (lowest resolution feature map), with no prior scoring, overlap $o$ is measured between all possible pairs to predict an overlap score map $O^{N \times N \times 1}$. Table \ref{tab:patch_comp} shows the reduction in patch comparisons at each scale for 256x256 image pairs. 

\subsection{Loss} We pretrain the endoder and overlap detection network jointly using the margin ranking loss function $\mathcal{L}_{o}$. The model is then trained end-to-end with mask output using binary cross-entropy loss. For two feature vectors $x_{1}$ and $x_{2}$ the regular margin ranking loss function is calculated as \eqref{eq:mrl}, where $m$ is the margin hyper-parameter. In our experiments for an anchor, positive and negative patch triplet $<a, a^{+}, a^{-}>$, $x_{1}$ and $x_{2}$ represent the overlap scores between patch pairs $<a, a^{+}>$ and $<a, a^{-}>$ respectively. Therefore the difference between $x_{1}$ and $x_{2}$ represents the difference in overlap between positive and negative patch pairs. As a result, we also experiment with a flexible margin that is measured as a function of overlap difference. Specifically, if the true overlap in pixels for $<a, a^{+}>$ and $<a, a^{-}>$ is $ o^{+}$ and $o^{-}$, the flexible margin $m_{flex}$ is shown in \eqref{eq:mflex}, where $d$ is the patch dimension. Then the flexible margin ranking loss $\mathcal{L}_{flex}$ is calculated as \eqref{eq:flex}.

\begin{equation}\label{eq:mrl}
    \mathcal{L}_{o} = max(0,(x_{2}-x_{1})+m)
\end{equation}

\begin{equation}\label{eq:mflex}
    m_{flex} = \frac{o^{+} - o^{-}}{d^{2}}
\end{equation}

\begin{equation}\label{eq:flex}
    \mathcal{L}_{flex} = max(0,(x_{2}-x_{1})+m_{flex})
\end{equation}

\subsection{Implementation and Training Details} We resize all input images to $256 \times 256 \times 3$ dimension. The largest scale has 8x8x256 dimension feature map for 32x32 dimensional patches. The channel dimension is halved at each scale (256 at scale 5 and 16 at scale 1). The overlap detection layer is a two layer feed-forward network. We pretrain our model for 25 epochs with the margin ranking loss on overlapping and non-overlapping patch triplets generated from synthetic data. The model is trained end-to-end for 50 epochs after that with binary cross-entropy and margin ranking loss. We use the adam optimizer with a learning rate of 1e-4.

\begin{figure*}[t!]
    \centering
    \includegraphics[width=0.85\linewidth]{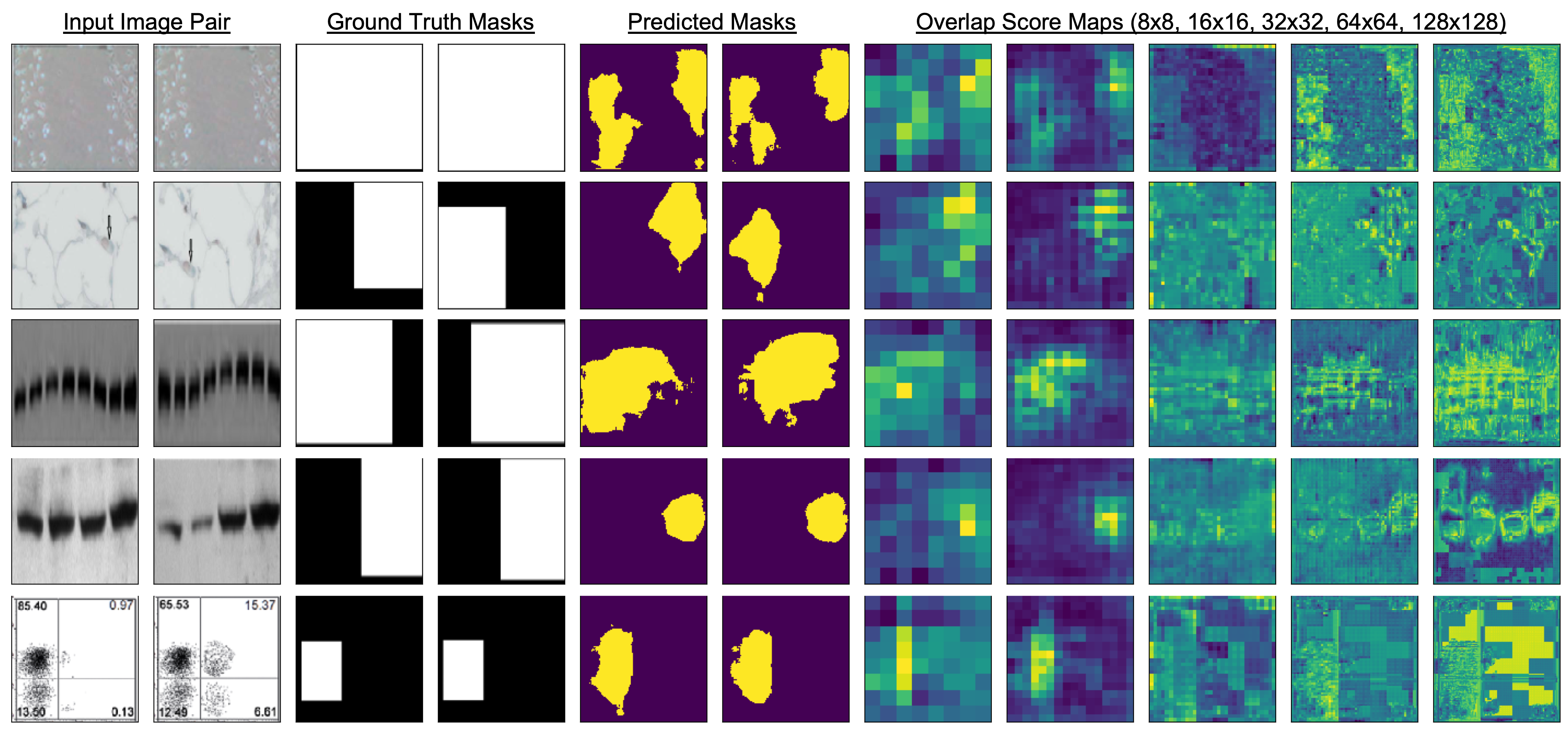}
    \caption{Input images, ground truth masks, predicted masks and intermediate score maps from MONet.}
    \label{fig:examples}
\end{figure*}
\section{Experiments}\label{sec:experiments}

\begin{table}[tbh]
\small
    \centering
    \begin{tabular}{ccc}
    \hline
         \textbf{Method} & \textbf{Image} & \textbf{Pixel} \\
         \cmidrule(lr){1-1}\cmidrule(lr){2-3}
         Ours w/o gating & 0.340 & 0.398 \\
         Ours w/ dot product overlap & 0.076 & 0.052 \\
         Ours - normalized margin & 0.398 & 0.410 \\
          Ours - flexible margin & 0.410 & 0.438 \\
         \hline
    \end{tabular}
    \caption{Ablation of gates or using dot-product for overlap.}
    \label{tab:ablation}
\end{table}

\subsection{Dataset and Metrics}

We use the BioFors dataset introduced in \cite{sabir2021biofors}. The EDD task has 1,547 manipulated images. Train and test splits have 30,536 and 17,269 images respectively, divided into four image categories -- Microscopy, Blot/Gel, Macroscopy and FACS. Each category has a different origin or semantic meaning, which leads to diverse image properties and challenges. We evaluate at the image and pixel level, according to the protocol described in \cite{sabir2021biofors}. Image level evaluation assigns binary labels to images. Pixel level evaluation is performed on aggregated pixel statistics across images. We use matthews correlation coefficient (MCC) metric as reported in \cite{sabir2021biofors}.

\subsection{Synthetic Data Generation}

BioFors dataset does not provide any manipulated samples for training. Hence, we train our model using synthetically generated samples similar to the process described in \cite{sabir2021biofors}. However, our model additionally requires joint pre-training of encoders and overlap scoring modules (OSMs). This requires extensive hierarchical annotation of patch overlap at each pixel i.e. patch pairs and their overlap scores at each scale. Generating such extensive annotation on the fly is computationally expensive. As a workaround, we generate predefined annotation templates, which can be used with random image-pairs on the fly to generate unique synthetic samples.

\subsection{Results}

Table \ref{tab:result} shows the performance of our model. Baseline results are presented as reported in \cite{sabir2021biofors}. Image and pixel columns denote corresponding evaluation protocol. We highlight two versions of our model -- with regular margin ranking loss and with a flexible margin ranking loss. Our model achieves a new state-of-the art on blot/gel, microscopy, macroscopy image categories and also on the combined evaluation. 

\begin{figure}[t!]
    \centering
    \includegraphics[width=0.85\linewidth]{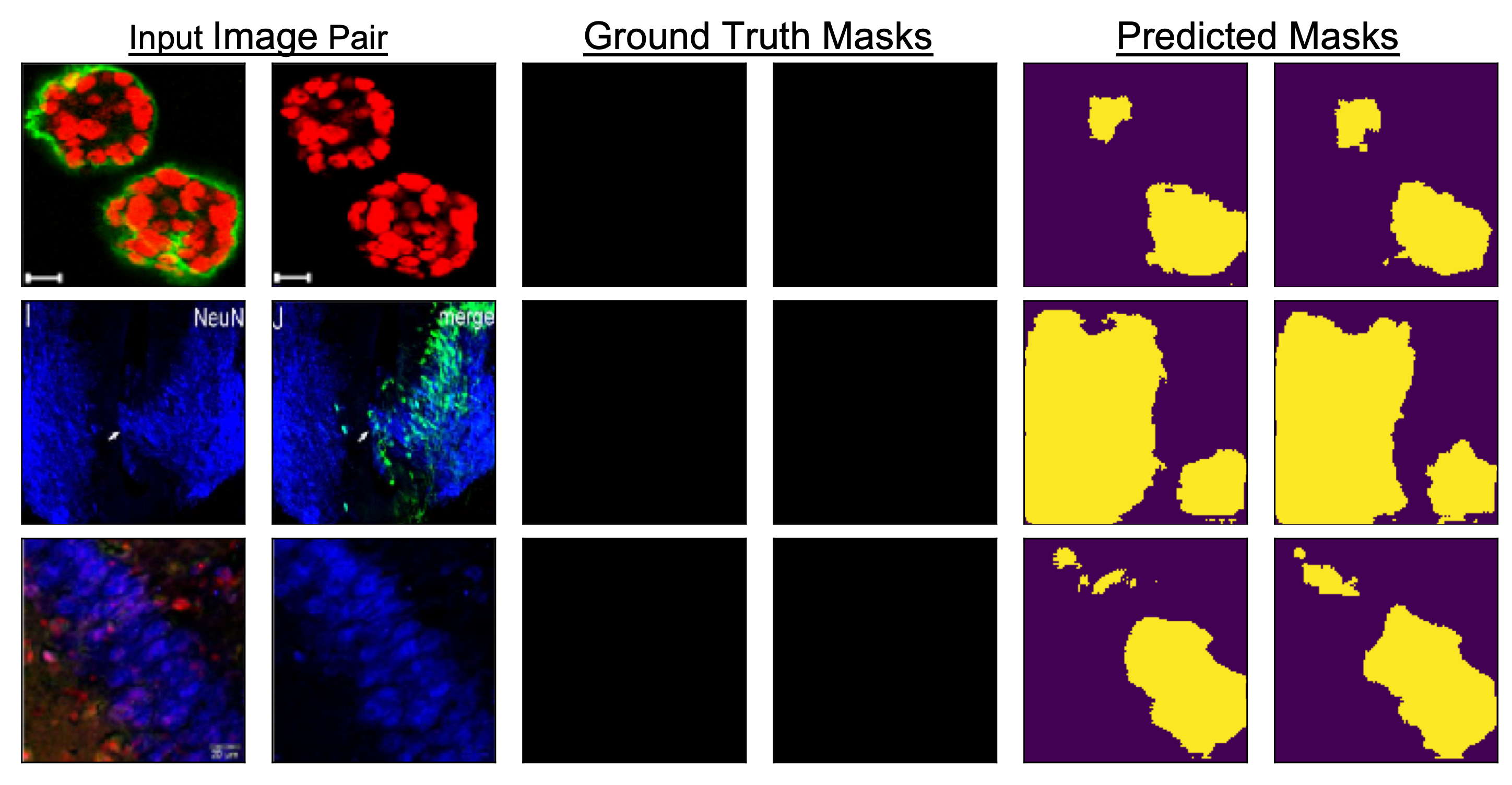}
    \caption{False positive samples on Microscopy images.}
    \label{fig:fp_staining}
\end{figure}

\subsection{Analysis}

As shown in Table \ref{tab:result}, our model achieves state-of-the art result across multiple categories. However, the performance fluctuates across image categories. Additionally, a single model does not hold top-performance across categories. We believe that the unique characteristics of each category make it difficult to train a single outperforming model. Figure \ref{fig:examples} shows sample predictions from our model. Overlap score maps from each scale show the progression of patch overlap detection. Figure \ref{fig:fp_staining} also shows that our method generates false positives. As described in \cite{sabir2021biofors}, these duplicated regions are not considered manipulations due to the semantics of experiments that produced them, such as image overlay or chemical staining. Overcoming these false positives requires either additional semantic information from source documents or the definition of manipulation needs rethinking.

\noindent\textbf{Ablation:} We perform an ablation analysis of our model in Table \ref{tab:ablation}. The model performance degrades if we remove the gating operation when concatenating overlap score maps across scales. Additionally, performance drops drastically if we use feature dot products from literature \cite{wu2017deep,Wu_2018_ECCV} instead of one hidden layer overlap detection network. 

\section{Conclusion}\label{conclusion}

Duplication of images across biomedical experiments is a concerning issue. Our proposed model achieves state-of-art-performance on the EDD task for some image categories. Further investigation with dedicated model for each image category is required.

\section{Acknowledgement}\label{acknowledgement}

This material is based on research sponsored by DARPA and Air Force Research Laboratory (AFRL) under agreement number FA8750-20-2-1004. The U.S. Government is authorized to reproduce and distribute reprints for Governmental purposes notwithstanding any copyright notation thereon. The views and conclusions contained herein are those of the authors and should not be interpreted as necessarily representing the official policies or endorsements, either expressed or implied, of DARPA and Air Force Research Laboratory (AFRL) or the U.S. Government.


\vfill\pagebreak
\clearpage
\balance
\bibliographystyle{IEEEbib}
\bibliography{refs}

\end{document}